\newcommand{\cv}[1]{}  %
\newcommand{\av}[1]{#1}    %

\av{
\documentclass[10pt,usletter]{article}
\usepackage{geometry}
\usepackage{amssymb} 
\usepackage{adjustbox}
\usepackage{hyperref}
\usepackage{mycitations}  %

\geometry{
    letterpaper,
    width=5.5in,
    height=8.5in,
    top=1in,
    headheight=14pt
}
} %

\usepackage{times}
\usepackage{soul}
\usepackage{url}

\usepackage[utf8]{inputenc}
\usepackage[small]{caption}
\usepackage{graphicx}
\usepackage{amsmath}
\usepackage{amsthm}
\usepackage{booktabs}
\usepackage{algorithm}
\usepackage{algorithmic}
\usepackage[switch]{lineno}
\usepackage{stfloats}
\usepackage{xspace}
\usepackage{tikz}
\usepackage{multirow}

\def\hy{\hbox{-}\nobreak\hskip0pt} 
\usepackage{boxedminipage}

\newcommand{\pbDef}[3]{%
    \noindent
    \begin{center}
        \begin{boxedminipage}{\cv{0.98}\av{0.75} \columnwidth}
            \textbf{Task:} #2 \\[2pt]
            \textbf{Encoding:} #3
        \end{boxedminipage}
    \end{center}
}

\newcommand{\QpbDef}[4]{%
    \noindent
    \begin{center}
        \begin{boxedminipage}{\cv{0.98}\av{0.75} \columnwidth}
            \textbf{Task:} #2 \\[2pt]
            \textbf{Encoding:} #3 \\[2pt]
            \textbf{Propagator:} #4
        \end{boxedminipage}
    \end{center}
}

\newcommand{\SMS}{SMS\xspace}
\newcommand{\GGG}{\mathcal{G}}
\newcommand{\n}[1]{\overline{#1}}
\newcommand{\SB}{\{\,}%
\newcommand{\SM}{\mid}
\newcommand{\SE}{\,\}}%
\newcommand{\var}{\text{var}}
\renewcommand{\mod}{\text{mod}}
\newcommand{\splitter}{\Gamma}
\newcommand{\rich}[1]{{#1}^{*}}
\newcommand{\scoringf}{\sigma}

\newcommand{\textcite}[1]{\citeauthor{#1}~[\shortcite{#1}]}

\cv{
\title{Smart Cubing for Graph Search: A Comparative Study}
\author{Anonymous Submission}
} %

\av{
\title{Smart Cubing for Graph Search: A Comparative Study%
\thanks{Research supported by the Austrian Science Fund (FWF)
within the projects  10.55776/36688 and
10.55776/COE12.}
}

\author{
Markus Kirchweger,
Hai Xia,
Tom\'{a}\v{s} Peitl, and
Stefan Szeider\\[4pt]
  \small  Algorithms and Complexity Group\\[-3pt]
  \small TU Wien, Vienna, Austria\\[-3pt]
\small [mk,hxia,peitl,sz]@ac.tuwien.ac.at
}
\date{}
} %

\begin{document}

\maketitle
\av{\thispagestyle{empty}}
\begin{abstract}
Parallel solving via cube-and-conquer is a key method for scaling SAT
solvers to hard instances. While cube-and-conquer has proven
successful for pure SAT problems, notably the Pythagorean triples
conjecture, its application to SAT solvers extended with propagators
presents unique challenges, as these propagators learn constraints
dynamically during the search.

We study this problem using SAT Modulo Symmetries (SMS) as our primary
test case, where a symmetry-breaking propagator reduces the search
space by learning constraints that eliminate isomorphic
graphs. Through extensive experimentation comprising over 10,000 CPU
hours, we systematically evaluate different cube-and-conquer variants
on three well-studied combinatorial problems. Our methodology combines
prerun phases to collect learned constraints, various cubing
strategies, and parameter tuning via algorithm configuration and
LLM-generated design suggestions.

The comprehensive empirical evaluation provides new insights into effective cubing strategies for propagator-based SAT solving, with our best method achieving speedups of 2-3x from improved cubing and parameter tuning, providing an additional 1.5-2x improvement on harder instances.

\end{abstract}

\section{Introduction}

Propositional satisfiability (SAT) solvers based on conflict-driven
clause learning can solve huge instances with millions of variables
and clauses~\cite{FichteLeberreHecherSzeider23}. However, for hard
instances, particularly in combinatorial problems, parallelization
becomes necessary. The cube-and-conquer technique has proven highly
effective for such problems, most notably in resolving the Pythagorean
triples conjecture~\cite{HeuleKullmannMarek16}.

In cube-and-conquer, a look-ahead solver first partitions the search space into disjoint sub-problems via cubes (partial assignments), which are then solved independently by CDCL solvers. This independence enables efficient parallel solving.

When encoding combinatorial problems into SAT, particularly those
involving graphs, we often encounter highly symmetric search
spaces. Many mutually isomorphic graphs satisfy the same constraints,
but a solver needs to check only one representative, the
\emph{canonical} element, from each isomorphism class. Standard CDCL
solvers cannot leverage these symmetries, and static symmetry breaking
methods cannot break all symmetries
\cite{CodishMillerProsserStuckey19}.

SAT Modulo Symmetries (SMS)
\cite{KirchwegerSzeider21,KirchwegerSzeider24} addresses this
limitation through dynamic symmetry breaking, using a custom
propagator that learns symmetry-breaking predicates during the
search. SMS has been successfully applied to several hard
combinatorial problems, including problems from extremal combinatorics
on on graphs, hypergraphs, matroids, and clause sets\footnote{\cite{FazekasNPKSB23,JanotaKirchwegerPeitlSzeider25,KirchwegerScheucherSzeider22,KirchwegerPeitlSzeider23,KirchwegerPeitlSzeider23b,KirchwegerScheucherSzeider23,ZhangPeitlSzeider24,ZhangSzeider23}}.

While one could apply standard cube-and-conquer to any SAT-based system with propagators, this approach yields poor performance. The key challenge is that propagators learn constraints dynamically which are not present in the initial formula. In SMS, for instance, the cubing solver working only with the input formula effectively partitions the space of all graphs rather than the space of canonical graphs. Other propagator-based systems face similar issues when the propagators contribute essential constraints to reduce the search space.

Our contribution is a systematic study of cube-and-conquer methods
adapted for SAT solving with propagators, using SMS as the primary
test case. Following a carefully designed experimental methodology, we
evaluate several cubing strategies and investigate effective
partitioning approaches for problems with dynamic constraints. Our
pipeline consists of a \emph{prerun phase} to collect important
learned clauses, \emph{look-ahead-based cubing} with various
heuristics, and a final \emph{solving phase}. Since cubes from the
same instance share structural properties, we use structured \emph{algorithm
  configuration}~\cite{SMAC3} to optimize the SAT solver specifically for the types of subproblems it will encounter. We also explore LLM-based design of look-ahead scoring functions that determine variable priorities during cubing. Our experimental evaluation comprises over 10,000 CPU hours across multiple benchmark problems.

Our \emph{results} demonstrate consistent improvements across our three benchmark problems: triangle-free graph coloring, Kochen-Specker graph enumeration, and diameter-2-critical graphs. The prerun and cubing phase alone provides a speedup between 2x and 3x compared to standard SMS. Parameter tuning  yields an additional 1.5x to 2x improvement, with the effect being most pronounced on harder instances. Our best-performing cubing strategy, based on established look-ahead techniques, consistently generates better-balanced subproblems than alternatives.

The remainder of the paper is organized as follows.
In Section~\ref{sec:prelims}, we review the notation used in the paper
and some common knowledge on graphs and propositional logic.
In Sections~\ref{sec:sms} and \ref{sec:cnc} thereafter, we introduce the two main characters in this paper: SMS and cube-and-conquer.
In Section~\ref{sec:sms-cubing}, we explain how we apply cube-and-conquer to SMS and give details of the various configurations.
In Section~\ref{sec:problems}, we provide a short background on our benchmark problems and encodings.
We present the results of our comparison in Section~\ref{sec:results}
and wrap up in Section~\ref{sec:conc}.

\av{\paragraph{Supplementary Material} Code and instances are available on
  Zenodo \cite{PeitlKSX25}.}

\section{Preliminaries}
\label{sec:prelims}

For a positive integer~$n$, we write $[n] = \{1,2,\dots,n\}$.
Below we review basics from propositional logic and graph theory.

\paragraph{SAT.}
A \emph{literal} is a propositional variable $x$ or its negation $\n{x}$.
A \emph{clause (cube)} is a disjunction (conjunction) of literals.
A \emph{CNF (DNF) formula} is a conjunction (disjunction) of clauses (cubes).
We sometimes interpret clauses/cubes as sets of literals and CNF (DNF) formulas as sets of clauses (cubes).
For a clause (cube) $C$, the cube (clause) $\SB \n{x} \SM x \in C \SE$ is denoted by $\n{C}$.
We write $\var(x) = \var(\n{x}) := x$ for the variable of a literal, $\var(C) = \SB \var(x) \SM x \in C$ for a clause/cube $C$, and $\var(F) = \bigcup_{C \in F} \var(C)$ for a CNF/DNF formula $F$.
An \emph{assignment} to a set of variables $V$ is a mapping $\tau : V \rightarrow \{ 0, 1 \}$, and is extended to literals by $\tau(\n{x}) = 1 - \tau(x)$.
An assignment $\tau$ \emph{satisfies} a clause $C$ if it maps at least one of its literals to~$1$.
An assignment to $\var(F)$ that satisfies every clause of $F$ is called a \emph{satisfying assignment} or a \emph{model}.
The set of models of a formula $F$ is denoted by $\mod(F)$.
If $\mod(F) \neq \emptyset$ we say that $F$ is \emph{satisfiable}, otherwise it is \emph{unsatisfiable}.
If every assignment $\tau : \var(F) \rightarrow \{ 0, 1 \}$ is a model, then $F$ is a \emph{tautology}.
For two formulas $F$ and $G$, if $\mod(F) \subseteq \mod(G)$, then $F$ \emph{entails} $G$, written $F \models G$.
Two formulas $F$ and $G$ are \emph{equisatisfiable} if they are both satisfiable or both unsatisfiable, and are \emph{logically equivalent} if $\mod(F) = \mod(G)$.
We will identify an assignment with the set of literals it maps to $1$.
For $V \subseteq \var(F)$ and an assignment to $V$ $\tau$, we write $F[\tau]$ for the formula obtained from $F$ by removing all literals mapped to $0$ by $\tau$, as well as all clauses satisfied by $\tau$.
A clause with only one literal is called a \emph{unit clause}.
If $F$ contains a unit clause $C = (x)$, then the literal $x$ must be set to $1$ in every model, and without loss of generality, we can focus on the simplified formula $F[x]$.
Repeated application of this rule until either there are no more unit clauses or an empty clause is created is called \emph{unit propagation}.

\paragraph{Graphs.}
We consider simple, undirected and unweighted graphs
without parallel edges or self-loops. A \emph{graph} $G$ consists of
a set $V(G)$ of vertices and a set $E(G) \subseteq \binom{V}{2}$ of edges; we denote the edge
between vertices $u,v\in V(G)$ by $uv$ or equivalently $vu$.
We write
$\GGG_n$ to denote the class of all graphs with $V(G) = [n]$.  The \emph{adjacency matrix} of a
graph $G \in \GGG_n$, denoted by $A_G$, is the $n\times n$ matrix where the element at row $v$ and column $u$, denoted by $A_G(v,u)$, is $1$ if $vu \in E$ and
$0$ otherwise. %
For a permutation $\pi : [n] \rightarrow [n]$, $\pi(G)$ denotes the graph obtained from $G\in \GGG_n$ by the
permutation $\pi$, where $V(\pi(G)) = V(G) = [n]$ and
$E(\pi(G))=\SB \pi(u)\pi(v)\SM uv \in E(G) \SE$.
Two graphs $G_1,G_2\in \GGG_n$ are \emph{isomorphic} if there is a
permutation $\pi : [n] \rightarrow [n]$ such that $\pi(G_1)=G_2$; in this case $G_2$ is an \emph{isomorphic copy} of $G_1$.

\section{SAT Modulo Symmetries (SMS)}
\label{sec:sms}

Modern propositional satisfiability (SAT) solvers are primarily based on \emph{conflict-driven clause learning (CDCL)}~\cite{FichteHLS23,MarquessilvaLynceMalik21}.
CDCL is a backtracking exhaustive search algorithm which decides the satisfiability of an input CNF as follows.
In a loop, a CDCL solver runs unit propagation until fixpoint (all forced assignments), then picks a variable to branch on when there is nothing left to propagate, and learns new clauses from conflicts obtained through unit propagation when they occur (when both $x$ and $\neg x$ is forced).
When a model is found, the solver reports it; when it learns the empty clause, it reports unsatisfiability.

SMS is a framework that augments a CDCL SAT solver%
\footnote{At the moment, SMS uses the SAT solver
 CaDiCaL~\cite{BiereFallerFazekasFleuryFroleyks24} and communicates
 with it through the IPASIR-UP
 interface~\cite{FazekasNPKSB24}.}
with a custom propagator that can reason about graph isomorphisms
(\emph{symmetries}), allowing the SAT solver to search \emph{modulo
 isomorphisms} for graphs with a given number $n$ of vertices and which satisfy constraints
specified by a propositional formula.
The SMS propagator is a routine which checks whether the currently explored branch of the search tree contains a \emph{canonical graph}.
A canonical graph is a distinguished member of its isomorphism class---in SMS, it is the isomorphic copy with a lexicographically minimal adjacency matrix, where matrices are compared as row-wise concatenated vectors: the vector of $\left( \begin{smallmatrix} 0 & 1 \\ 0 & 0 \end{smallmatrix} \right) $ is $\left( \begin{smallmatrix} 0 & 1 & 0 & 0 \end{smallmatrix} \right)$.
This routine is thus referred to as the \emph{minimality check}.
The minimality check is called by the CDCL solver after unit propagation reaches fixpoint, and can trigger an additional conflict on
top of ordinary CDCL and consequently learn a \emph{symmetry-breaking clause}, which is a clause that excludes non-canonical graphs.

In order for SMS to check minimality, a graph from the current partial assignment of the CDCL solver has to be constructed.
This is done by looking at the values of the specially designated \emph{edge variables}, which, by convention, are the first $\binom{n}{2}$ variables of the formula (when searching for graphs with $n$ vertices; the value $n$ must be passed to SMS).
These variables map to the upper triangle of the adjacency matrix in
row-major order
$\left( \begin{smallmatrix} \cdot & 1 & 2 & 3 \\ \cdot & \cdot & 4
    & 5 \\ \cdot & \cdot & \cdot & 6 \\ \cdot & \cdot & \cdot & \cdot \end{smallmatrix} \right)$,
and from their assignment one can extract a \emph{partially defined graph}, which is a graph in which the presence of some edges is unknown (whose variables are unassigned).
SMS learns a symmetry-breaking clause when the current partially defined graph cannot be extended to any canonical graph.

The minimality check is only one of a set of external propagators implemented in SMS.
These propagators can be used as a substitute for constraints that would be difficult to encode in propositional logic.
A typical such example is non-3-colorability, which cannot be encoded in a polynomial-size propositional formula unless $\mathrm{NP}=\mathrm{coNP}$; more on the use of propagators in SMS in Section~\ref{sec:problems}.
Lexicographic minimality of the adjacency matrix among isomorphic copies is, like non-3-colorability, also a coNP-complete property~\cite{CrawfordGinsbergLuksRoy96}.

\citeauthor{KirchwegerSzeider24}~\shortcite{KirchwegerSzeider21,KirchwegerSzeider24} give a comprehensive description of SMS.
For the purpose of this paper, the salient property of SMS is that parts of the encoding---namely, the symmetry-breaking clauses and any clauses added by other propagators---are learned dynamically during the solver run and are unknown at the beginning.
As a consequence, we cannot simply reuse existing SAT parallelization methods, as those assume (of course) that the entire problem is represented in the encoding.
In Section~\ref{sec:sms-cubing}, we explain how to take the external propagators into account properly for parallelization; before that, we review the SAT parallelization technique cube-and-conquer.

\section{Cube-and-Conquer}
\label{sec:cnc}

Assume that a hard CNF formula $F$ is to be solved by a SAT solver, and let $x$ be a variable of $F$.
Consider the formulas $F[x]$ and $F[\n{x}]$ where $x$ is assigned true or false, respectively.
Clearly, $F$ is satisfiable if, and only if, at least one of $F[x]$ and $F[\n{x}]$ is satisfiable.
Therefore, instead of solving~$F$, one can solve $F[x]$ and $F[\n{x}]$, the point being that these \emph{sub-problems} can be solved in parallel, and will hopefully be easier to solve than the original formula $F$.
This reasoning applies equally for enumeration problems: in order to enumerate the models of $F$, one can separately enumerate the models of $F[x]$ and $F[\n{x}]$, and take the union of the two.

This basic splitting idea can, of course, be applied recursively: by re-splitting the sub-problem $F[\n{x}]$ we obtain the collection of sub-problems $F[x]$, $F[\n{x},y]$, and $F[\n{x},\n{y}]$.
In general, given an input formula $F$, and a DNF tautology $\splitter = C_1 \lor \dots \lor C_r$ over (some of) the variables of $F$, the set of models of $F$ is obtained as $ \mod(F) = \bigcup_{i=1}^r \mod \left( F[C_r] \right) $.
The $C_i$ are called \emph{cubes}, and the method that solves $F$ by designing a suitable cube set $\splitter$ to divide into sub-problems and solving each sub-problem separately (typically in parallel) is called \emph{cube-and-conquer}~\cite{HeuleKullmannMarek16,HeuleKullmannBiere18}.

Note that $\splitter$ does not have to be a tautology.
For satisfiability checking in ordinary SAT (no SMS or other special propagators), it is sufficient that $F \land \splitter$ is equisatisfiable to $F$.
In our context, we add symmetry-breaking clauses on the fly, and we enumerate models instead of just checking satisfiability. We thus must require preservation of all models, i.e., that $F \models \splitter$, or, equivalently, that $F \land \lnot \splitter$ is unsatisfiable.

A fundamental optimization problem emerges: how to find a good splitter $\splitter$ in order to optimize solve (conquer) performance?
The standard answer to this question is \emph{look-ahead} solving.
A look-ahead solver constructs a branching tree whose nodes are variables.
At every node, the solver tries assigning each as yet unassigned variable in both possible ways and performs the associated formula reduction (repeated removal of falsified literals and satisfied clauses followed by unit propagation).
It then collects information about the reduced formula in each branch into a numerical value and combines both values into a variable score using a \emph{scoring function}.
The goal of the scoring function is to produce a single number on which all variables can be compared, which reflects both the total amount of reduction resulting from the two assignments as well as how balanced the reduction is between the two branches.
The variable that scores highest on this aggregate score is picked as the next branching variable, and two subproblems are created, on which the solver can be called recursively until a cutoff threshold is reached (typically specified as the number of variables to be assigned in total).

This basic method could be directly applied to SMS, but it is not entirely suitable for its own use.
This is because, in SMS, not all information about the problem is encoded upfront in the formula~$F$.
Part of the encoding is ``learned'' dynamically, such as the symmetry-breaking clauses or other clauses added by propagators in SMS.
In Section~\ref{sec:sms-cubing}, we explain a cube-and-conquer adaptation that is tailored to SMS and scenarios with custom propagators in general.

\section{Cubing with SMS}
\label{sec:sms-cubing}

The main contribution of this paper is the design and experimental evaluation of various cube-and-conquer pipelines for SMS.
We start with an encoder that produces a propositional CNF $F$.
In traditional cube-and-conquer, one would start cubing $F$ directly, but in our scenario, this would result in poor performance.
Instead, we start with the \emph{prerun} phase, in which $F$ is enriched with additional clauses to obtain a formula $\rich{F}$.

During the prerun, we run SMS, i.e., CaDiCaL and its attached propagators, for a bounded amount of time.
The point of this prerun phase is to
improve the performance of the cubing phase by
collecting a set $\Sigma$ of
symmetry-breaking clauses, a set $\Pi$ of propagator clauses  as well
as a set  $\Lambda$ of important learned clauses from CDCL.
The most important learned clauses for $\Lambda$ are unit clauses, but we also store and reuse learned clauses with up to 5 literals.
We may find models during prerun; we simply store them and block in $\rich{F}$.
The enriched formula $\rich{F}$ is obtained as $F \land \Sigma \land \Pi \land \Lambda$.

After the prerun, we save the enriched CNF $\rich{F}$ and pass it on to the cuber.
The cuber produces a set $\splitter$ of cubes to be solved by the final-phase solver.

The final solver is the same one as the one used for the prerun, up to the configuration of parameters.

Within this general framework, we evaluate a number of possible choices for each of the components and for the ways the components interact.

\subsection{Prerun}
For prerun, we always run CaDiCaL with SMS, and with a suitable domain-specific propagator as necessary for a benchmark problem (see Section~\ref{sec:problems} for details on the benchmark problems and propagators).
When the prerun phase is followed by a cubing phase that uses CaDiCaL, the solver simply switches from prerun to cubing mode, and continues to run, preserving all learned clauses and any other acquired state.
When cubing is done externally, the solver dumps $\Sigma$, $\Pi$, and $\Lambda$ into a file and stops.

\subsection{Cuber}
For the cubing component, we evaluate several different methods:

\newcommand{\propname}{Propagator}
\paragraph{CDCL with a cutoff.}
In this method, we continue running SMS as in prerun, but whenever the
number of assigned edge variables reaches a predefined threshold, we
collect all assigned edge literals, output them as a cube, and add the
negation as an irredundant learned clause.
Thus, the generation of cubes blindly follows whatever path the CDCL solver takes.
This method has been fruitfully employed in previous applications of   SMS~\cite{KirchwegerPeitlSzeider23}.
When using this method, we run SMS only once: after the amount of time given for prerun has elapsed, the solver simply switches to cubing mode and continues to run.
Note that this also means that the resulting cubes highly depend on the solver state at the end of prerun, meaning that different runs can result in very different cubes. 
The process can be depicted in this diagram.
\av{\begin{center}}
\begin{tikzpicture}
	\pgfmathsetmacro{\solx}{6}
	\tikzstyle{propnode}=[draw, rounded corners=1]
	\tikzstyle{mainnode}=[draw]
	\node[style=mainnode] (enc) at (0,0) {Encode};
	\node[style=mainnode] (pre) at (3,0) {Prerun-then-Cube};
	\node[style=propnode] (ppr) at (3,1) {\propname};
	\node[style=mainnode] (sol) at (\solx,0) {Solve};
	\node[style=propnode] (spr) at (\solx,1) {\propname};

	\draw[-latex] (enc) -- node [above,midway] {$F$} (pre);
	\draw[-latex] (pre) -- node [above,midway] {$F \land \splitter$} (sol);

	\draw[-latex, solid]  (pre) to [bend left=20] (ppr);
	\draw[-latex, solid]  (ppr) to [bend left=20] (pre);

	\draw[-latex, solid]  (sol) to [bend left=20] (spr);
	\draw[-latex, solid]  (spr) to [bend left=20] (sol);
\end{tikzpicture}
\av{\end{center}}
\paragraph{CaDiCaL with look-ahead.}
In this method, we simulate a look-ahead solver within CaDiCaL (inside SMS).
We control the assignments made by CaDiCaL through the IPASIR-UP interface, using the \texttt{force\_backtrack} function to navigate the search tree.
At each decision point, we try to assign each candidate variable in both ways, counting the number of propagations $a$ and $b$ triggered in the two branches. If either assignment results in a conflict, we assign the variable to the opposite truth value and start the lookahead again. Otherwise, we backtrack one decision level and test another variable. 
If all variables are tested, we pick the variable for which the total number of propagations is as high and as balanced as possible.
For this we use a \emph{scoring function} $\scoringf$, which balances these two aspects, and is a parameter of the method.
The default scoring function is $\scoringf(a, b) = \min(a, b) +
\varepsilon(a+b)$, where $\varepsilon = 10^{-9}$ is a small positive number
The set of candidate variables can either be all unassigned variables, or all unassigned edge variables.
The rationale for the latter choice is that it is cheaper to look ahead on a smaller set of variables.
The structure of the process is the same as before, with the only difference that the `Cube` phase in `Prerun-then-Cube` is performed differently.

\paragraph{March\_cu.}
March\_cu is a genuine look-ahead solver with a long history of evolution~\cite{HeuleDZM04,HeuleM06,HeuleM09}.
It works similarly to the previous method, except it takes into account more characteristics of the formula than just the number of propagations in each branch. Another difference is that the cubes it produces contain only branching literals. Therefore, the cubes do not contain information about all assigned variables including variables assigned by unit clauses.
The scoring function used by March\_cu is $\sigma(a, b) = a + b + ab$.
The process looks as follows.

\av{\begin{center}}
\begin{tikzpicture}
	\pgfmathsetmacro{\solx}{6}
	\tikzstyle{propnode}=[draw, rounded corners=1]
	\tikzstyle{mainnode}=[draw]
	\node[style=mainnode] (enc) at (0,0) {Encode};
	\node[style=mainnode] (pre) at (2,0) {Prerun};
	\node[style=propnode] (ppr) at (2,1) {\propname};
	\node[style=mainnode] (cub) at (4,0) {Cube};
	\node[style=mainnode] (sol) at (\solx,0) {Solve};
	\node[style=propnode] (spr) at (\solx,1) {\propname};

	\draw[-latex] (enc) -- node [above,midway] {$F$} (pre);
	\draw[-latex] (pre) -- node [above,midway] {$\rich{F}$} (cub);
	\draw[-latex] (cub) -- node [above,midway] {$\splitter$} (sol);

	\draw[-latex, solid]  (pre) to [bend left=20] (ppr);
	\draw[-latex, solid]  (ppr) to [bend left=20] (pre);

	\draw[-latex, solid]  (sol) to [bend left=20] (spr);
	\draw[-latex, solid]  (spr) to [bend left=20] (sol);

	\draw[-latex, dashed] (pre) to [bend left=36] node [above,midway] {$F \land \Lambda$} (sol);
\end{tikzpicture}
\av{\end{center}}

For CaDiCaL with look-ahead, we additionally configure the scoring function $\scoringf$.
As a reference point we start with the default function $\scoringf_0(a, b) =  \min(a, b) + \varepsilon(a + b)$, where $a$ and $b$ are the number of propagations triggered in the two branches.
We run five rounds of the following procedure.
We take the best-performing scoring function so far, and query OpenAI GPT-4o (via the ChatGPT web interface) to suggest four variants of the function that could potentially work better.
For this, we informally describe the problem setting and the goal of the design to the LLM.
We then evaluate each of the scoring functions on a training set of cubes (for details see Section~\ref{sec:results}), and repeat the process with the winner.

\subsection{Conquering solver}
For the final solver, we take CaDiCaL again, but this time we configure its parameters with the state-of-the-art automatic algorithm configuration tool SMAC~\cite{SMAC3}.
The final solver is thus the same one as used for prerun, up to possibly changed search strategies. However, there are more than 100 various
adjustable strategies according to the exposed parameter list of CaDiCaL, and it is quite challenging to configure them together. Therefore,
we first run a sensitivity analysis to identify promising parameters for configuring: we switch each parameter individually to an extreme value and compare the changed configuration 
to the default configuration on a set of cubes sampled from a fixed problem. Based on this, we identify a small set of decisive parameters that we then configure with SMAC.
For details of the automatic configuration process, see Section~\ref{sec:results}.

All of the above described cubing configurations are summarized in Table~\ref{table:configs}.

\begin{table}
	\centering
	\begin{tabular}{lcc}
		\toprule
		    cuber                & LA scope & $\scoringf$ \\
		\midrule
			\texttt{CDCL}        &       NA &          NA \\
			\texttt{CaDiCaL-LA}  &     full &         any \\
			\texttt{CaDiCaL-LA}  &     edge &         any \\
			\texttt{march-cu}    &     full &     default \\
		\bottomrule
	\end{tabular}
	\caption{A summary of cubing configurations.
		LA stands for look-ahead.
		We tested CaDiCaL looking ahead on either all or just the edge variables.
		The values ``any'' mean that any $\scoringf$ can be used by these cubers in principle; in practice we optimized $\scoringf$ using suggestions from OpenAI GPT-4o only with the full-scope CaDiCaL-LA cuber, and used the best result with the other variant as well.
	}
	\label{table:configs}
\end{table}

\section{Benchmark Problems and Encoding}
\label{sec:problems}

In this section, we introduce the graph search problems and present
encodings on which we evaluate our methods.
For most problems, we only sketch the encoding \emph{and refer to the
  original source}, as our main focus is on comparing the solving approaches.
See Section~\ref{sec:results} for a link to generator scripts and details of the formulas.

Our benchmark set consists of three different problems.
In the first two cases (Subsections~\ref{sec:triangle-free-def} and
\ref{sec:ks-def}), the problems cannot be efficiently encoded into
propositional logic, and external propagators (already implemented and available in
the SMS library) are required.
For these two problems, we describe what the propagator does.
The last problem (Subsection~\ref{sec:diameter}) is encoded fully in propositional logic.

On top of the individual encoding (and propagators if necessary) for each problem, we also use incomplete static symmetry breaking constraints proposed by \textcite{CodishMillerProsserStuckey19}.
These constraints are compatible with SMS in the sense that they are satisfied by the lexicographically minimal graph, but they are incomplete because they are satisfied by non-minimal graphs as well.
We use these constraints to increase the amount of information about graph minimality in the formula (in addition to symmetry-breaking clauses found during the prerun phase).

\subsection{Coloring Triangle-Free Graphs}
\label{sec:triangle-free-def}

A \emph{proper $k$-coloring} of a graph $G$ is a mapping
$c : V(G) \to \{1,\dots,k\}$ such that $uv \in E(G)$ implies $c(u) \neq c(v)$.
The \emph{chromatic number} of a graph $G$ is the smallest integer
$k$, for which a proper $k$-coloring exists.
If a graph contains a complete subgraph on $k$ vertices, then its chromatic number must clearly be at least $k$.
The opposite is not true: \textcite{Mycielski55} explicitly
constructed triangle-free graphs (without triangle subgraphs) with
unbounded chromatic number.
Erd\H{o}s~\shortcite{Erdos67} asked about the values $f(k)$, which denote the smallest number of vertices in a triangle-free non-$(k-1)$\hy colorable graph.
Mycielski's construction provides %
upper bounds on $f(k)$, and these are tight up to $k=4$; for $k=5$, minimal graphs are also known~\cite{Goedgebeur20}, but none of them is a \emph{Mycielskian}.
The cases $k \geq 6$ are open.

\newcommand{\col}{c}

\QpbDef{(Max\hy)$\triangle$\hy free non-$k$-colorable}
{Compute a triangle-free graph with $n$ vertices and chromatic number at least $k$. }
{We enumerate all triples of vertices to ensure triangle-freeness. 
Without loss of generality, we further restrict the search to  \emph{maximal triangle-free} graphs (triangle-free and adding any edge creates a triangle).}
{When a canonical propositional model is found, the propagator checks whether it
  is $k$-colorable. If so, a \emph{coloring clause} is learned, which ensures that this particular coloring will not work for future candidate graphs. For details we refer to~\cite{KirchwegerPeitlSzeider23}.}

\subsection{Kochen-Specker Graphs}
\label{sec:ks-def}

\emph{Kochen-Specker (KS) vector systems} are special sets of vectors in at least
3-dimensional space that form the basis of the Bell-Kochen-Specker
Theorem, demonstrating quantum mechanics' conflict with classical
models due to
contextuality~\cite{BudroniEtal22}. \textcite{KochenSpecker67}  originally
came up with a 3D KS vector system of size~117. The smallest
known system (in 3D) has 31 vectors~\cite{Peres91}, while
the best lower bound is 24 \cite{KirchwegerPeitlSzeider23,LiBrightGanesh24}.  
These lower bounds were obtained with computer search for \emph{KS
candidate graphs}: \emph{non-010-colorable} graphs with additional restrictions.
A graph is 010-colorable if its vertices can be colored red and blue such that no two adjacent vertices are both red and no triangle is all blue.

\QpbDef{Kochen-Specker graphs}
{Enumerate all KS candidates with $n$ vertices.}
{For the full list of constraints and the encoding, see~\cite{KirchwegerPeitlSzeider23}.}
{When a canonical model is found, the propagator checks whether it is 010-colorable, and if so learns a \emph{coloring clause}, which ensures that at least one edge or triangle must be present to invalidate this particular coloring. For details we refer to~\cite{KirchwegerPeitlSzeider23}.}

\subsection{Diameter-2-Critical Graphs}
\label{sec:diameter}
The \emph{diameter} of a graph $G$ is the largest distance between
a pair of vertices in $G$, where the \emph{distance} of two vertices is
the length of a shortest path between them. A disconnected graph has
diameter $\infty$.  A graph is \emph{diameter\hy $d$\hy critical} if
its diameter is $d$ and deletion of any edge increases the
diameter. The study of extremal properties of graphs with prescribed
diameter was initiated by Erd\H{o}s and R\'{e}nyi~\shortcite{ErdosRenyi62} and has been the subject of intensive
research.  An important topic in the field is the characterization of
diameter\hy $d$\hy critical
graphs~\cite{ChenFuredy05,HaynesEtal15,LohMa16};
in particular the case $d=2$.
The \emph{Murty-Simon Conjecture}
\cite{CaccettaHaggkvist79} states that for a diameter\hy
$2$\hy critical graph with $n$ vertices and $m$ edges,
$m \leq \lfloor n^2 /4 \rfloor$, with equality attained only by the
complete bipartite graph
$K_{ \lceil n/2 \rceil, \lfloor n/2 \rfloor }$ (the related classic theorem of \textcite{Mantel07} postulates the same for $C_3$-free graphs).

\av{\sloppypar}Using Nauty~\cite{McKayPiperno14}, \textcite{RadosavljevicZivkovic20} computed
all diameter\hy $2$\hy critical graphs with up to 10 vertices.
\textcite{DaillyFoucaudHansberg19} reported on a ``computer
search'' for graphs with up to 11 vertices, focusing on graphs with a
certain number of edges. \textcite{KirchwegerSzeider24} enumerated all diameter-$2$-critical graphs up to $13$ vertices with \SMS.

We use the SAT encoding from this last work---a
CNF formula $D_2(n,m)$ which is satisfiable if and only if there is a
diameter\hy 2\hy critical graph $G$ with $n$ vertices and $m$
edges.
By a theorem of \textcite{Fan87}, the bound holds for all diameter-2-critical graphs with up to 24 vertices.
We therefore set $m = \lfloor n^2 / 4 \rfloor$ to enumerate diamater-2-critical graphs that attain equality.

\pbDef{Murty-Simon conjecture}
{Enumerate diameter-2-critical graphs with $n$ vertices and $m$ edges, in particular for $m = \lfloor n^2 / 4 \rfloor$.}
{We use variables $c_{i,j,k}$ to indicate whether $k$ is a common neighbor of $i$ and $j$, i.e., $ c_{i,j,k} \leftrightarrow (e_{i,k} \land  e_{j,k})$.
With these variables it is easy to encode diamater-2-criticality; for
details see the  article  by \textcite{KirchwegerSzeider24}.}

\section{Results}
\label{sec:results}
\newcommand{\default}{\texttt{Def}}
\newcommand{\tuned}{\texttt{Tun}}

In this section, we will present the technical details of our experiments
and the results. The overarching goal is to minimize the time taken
for the entire pipeline, from encoding to obtaining the solutions.
However, since the entire setup is quite complicated, we decided to
simplify some aspects.

We optimize the cubing and the conquering solver separately from each other.
For each benchmark problem, we create a training and a test instance.
The training instance is an easier instance of the problem with a smaller number of vertices, and the test instance is a harder instance with one more vertex.
While this might not sound like a big step, the search space generally grows exponentially in the number of edges, so roughly like $\exp(\Omega(n^2))$.
We want to note that for our selected combinatorial problems, the increment of the number of vertices from  $n$ to $n+1$ typically 
makes the problem by orders of magnitude harder. Sometimes, in these problems, the solution is known for a particular $n$ but not for $n+1$ so far.
For Kochen-Specker and triangle-free graphs, we train on 21 and test on 22 vertices; for the Murty-Simon conjecture, we train on 15 and test on 16 vertices.
Details on how to produce the encodings and cubes can be found in the supplementary material.

\paragraph{Technical parameters.}
We use CaDiCaL v2.1.2.
When in look-ahead mode, we turn off restarts and non-chronological backtracking.
We use a forked version of March\_cu obtained from \url{https://github.com/BrianLi009/MathCheck/tree/main/gen_cubes/march_cu}.
We ran into memory problems when attempting to run the version of March\_cu from \url{https://github.com/marijnheule/CnC}.
We ran the cubing phase of the experiments on a cluster of machines equipped with 16× Intel Xeon E5-2640 v4, 2.40GHz 10-core processors and 160GB of RAM, running Ubuntu 18.04.
We ran the conquering phase on a cluster equipped with 3× AMD EPYC 7402, 2.80GHz 24-core processors and 1024GB of RAM, running Ubuntu 18.04.

\subsection{SMS Configuration}

Let us for a moment fix the cubing component at the beginning of the default cuber \texttt{CDCL}.
Our goal is to optimize the search strategies of the conquering solver.
We first list all parameters of CaDiCaL, and pick those that are relevant to solver performance.
We run a sensitivity analysis for each parameter.
This means we take the default configuration, and for each parameter, we toggle it to a different value.
In the case of numerical parameters, we change the value to the extreme end of the admissible range.
We evaluate each thus obtained configuration on the set of cubes produced with the default cuber on each of the three training instances.
With this, we identify a set of 10 parameters that seem promising for further automated tuning with SMAC. Along with the other two parameters
for SMS itself (`\texttt{frequency}' and `\texttt{cutoff}'), the total 12 parameters are listed in the supplementary material.

In Figure~\ref{fig_parameters_ranking}, we list the best-ranked  CaDiCaL parameters among sets of problems (cubes) with different default solving times. 
Each colored line indicates the rankings of a parameter on different sets of instances with specific default solving time ranges. We can clearly see that the three best-ranking
parameters seldom drop out of the top league. For example, the parameter `\texttt{BRD}' is not only the most influential parameter when solving the cubes
with default running time between 60 and 960 seconds but also the second most decisive parameter among the cubes with longer default solving time.
\begin{figure}
	\centering
\cv{	\includegraphics[scale=0.4,trim=0 80 0
  100]{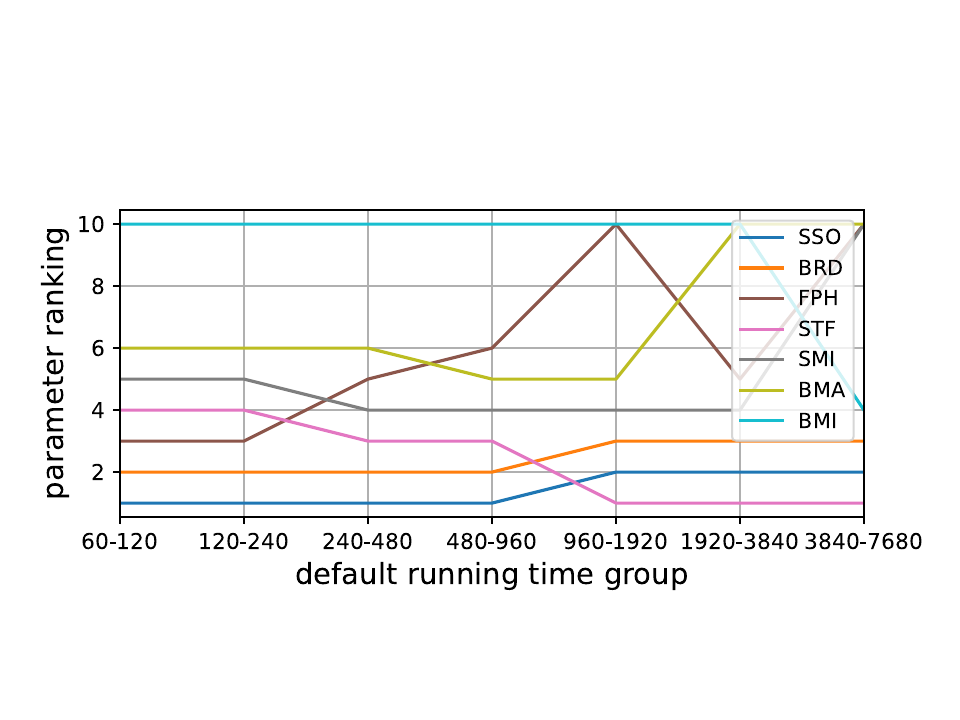}}
\av{	\includegraphics[scale=0.6,trim=0 80 0
  100]{config_ranking_plot_aspect_ratio.pdf}}

\av{\medskip}
\caption{The parameters ranking of CaDiCaL parameters. $10$ means a very poor rank, and that is the reason why different parameters can sometimes get the same rank.}
	\label{fig_parameters_ranking}
\cv{	\vspace{-2ex}}
\end{figure}

We ran such sensitivity analysis for each cuber, but the significant parameters were similar in each case, and for efficiency reasons 
we decided to proceed with the SMAC phase only on the cubes from the default cuber.
From this run of SMAC, we obtain, for each benchmark problem, a configuration of the parameters of SMS tuned specifically for this problem.
Below, we refer to this obtained configuration simply as \tuned, and to the default configuration as \default\ (the default is the same for each benchmark 
problem, but ``tuned'' is different for each problem).

\subsection{Configuration of $\sigma$}

We optimize the scoring function $\sigma$ through an iterative process using GPT-o1. As explained in Section~\ref{sec:sms-cubing},
starting with the default function, we run five rounds where
GPT-o1 is asked to generate four variant scoring functions of the current performing one.
The new functions are then evaluated by solving the cubes (generated by the full-variable look-ahead with the new functions) with the CaDiCaL with default configurations. This process yields specialized scoring functions for each benchmark problem whose intricate balance of terms defies intuitive human interpretation, 
yet demonstrates improved performance, as shown in the below.
Finally, we pick the best-performing scoring functions among training
instances for testing instances (harder instances), here are the top three.
\begin{align*}
\scoringf_{\mathrm{KS}}(a,b)
&=
8 \min(a,b)
\;+\;
2 \,\frac{\min(a,b)}{\max(a,b)+1} + a + b\\
	\scoringf_{\mathrm{TF}}(a,b) &= \min(a,b) \;+\; 10 \, \Bigl(\tfrac{\min(a,b)}{\max(a,b) + 1}\Bigr)^{2}\\
	\scoringf_{\mathrm{SMC}}(a,b) &=\; ab + a + b.
\end{align*}

\subsection{Test performance}

\begin{figure*}[ht]
  \centering
  \cv{
		\includegraphics[width=0.9\textwidth]{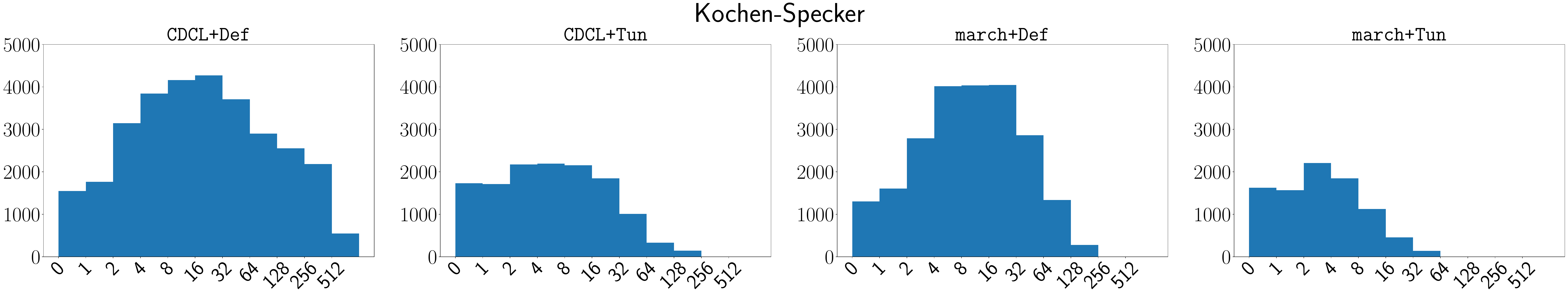}
		\includegraphics[width=0.9\textwidth]{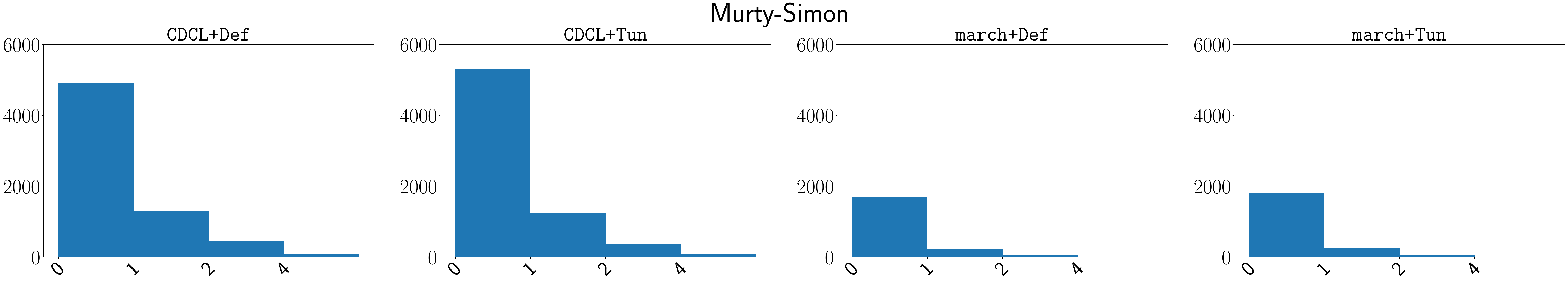}
		\includegraphics[width=0.9\textwidth]{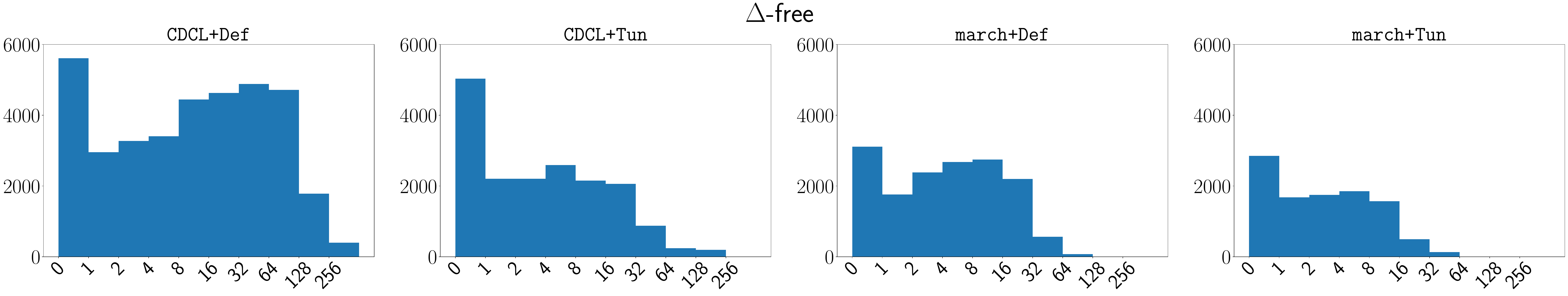}}
  \av{
		\includegraphics[width=\textwidth]{ks.pdf}
		\includegraphics[width=\textwidth]{msc.pdf}
		\includegraphics[width=\textwidth]{tf.pdf}}

	\caption{
		The total time to solve all cubes grouped by cube hardness.
		For each benchmark we compare default and march-based cubing and \default\ and \tuned\ conquering solver.
		The height of a bar between $x_0$ and $x_1$ is the total amount of time needed to solve the cubes which took between $x_0$ and $x_1$ minutes.
		The rightmost bar collects all remaining cubes.
		The total area of each chart is proportional to the total solving time reported in Table~\ref{table:default-vs-tuned};
		all four charts for one problem use the same unit of area, and are thus pairwise comparable.
	}
	\label{fig:histo}
	\vspace{-2ex}
\end{figure*}

We compare different pipelines on the total solving time for all cubes, or in other words on the time that it would take to solve all cubes on a single processor.
Since cubes are typically to be solved in parallel, we also report the time taken to solve the hardest cube for each pipeline, but this is not the main metric.
This information is shown in Table~\ref{table:default-vs-tuned}, for all cubers and for \default\ and \tuned, on the test instances.

\av{
\begin{table*}[t]
  \centering
  \begin{adjustbox}{width=\linewidth}
\begin{tabular}{@{}lrrrrrrrrrrrr@{}}
\toprule
cuber & \multicolumn{4}{c}{Kochen-Specker} & \multicolumn{4}{c}{Murty-Simon} & \multicolumn{4}{c}{$\Delta$-free} \\
      & \multicolumn{2}{c}{sum} & \multicolumn{2}{c}{$\max$} & \multicolumn{2}{c}{sum} & \multicolumn{2}{c}{$\max$} & \multicolumn{2}{c}{sum} & \multicolumn{2}{c}{$\max$} \\
      & \default & \tuned & \default & \tuned & \default & \tuned & \default & \tuned & \default & \tuned & \default & \tuned \\
\midrule
\texttt{CDCL} & $510.4$ & $221.7$ & $9.1$ & $2.4$ & $112.4$ & $116.8$ & $0.1$ & $0.1$ & $601.6$ & $292.8$ & $6.6$ & $3.3$ \\
\texttt{CaDiCaL-LA-E} & $836.0$ & $290.3$ & $17.7$ & $7.9$ & $60.8$ & $56.0$ & $0.2$ & $0.2$ & $657.1$ & $260.3$ & $16.9$ & $7.3$ \\
\texttt{CaDiCaL-LA-}$\forall$ & $858.4$ & $296.5$ & $17.4$ & $3.4$ & $78.3$ & $71.7$ & $0.3$ & $0.2$ & $570.7$ & $219.7$ & $16.3$ & $5.4$ \\
\texttt{march} & $\mathbf{371.2}$ & $\mathbf{149.4}$ & $\mathbf{2.5}$ & $\mathbf{0.9}$ & $\mathbf{33.5}$ & $\mathbf{35.6}$ & $\mathbf{0.1}$ & $\mathbf{0.1}$ & $\mathbf{258.6}$ & $\mathbf{172.2}$ & $\mathbf{1.2}$ & $\mathbf{0.9}$ \\
\bottomrule
\end{tabular}
  \end{adjustbox}
  \caption{Total solving time (sum) and time to solve the hardest cube $\left( \max \right)$ in CPU hours for cubes produced for test instances by different cubers (rows) and solved by default or hyper-parameter-tuned SMS. The best cuber for each column is in boldface.}
  \label{table:default-vs-tuned}

\end{table*}
}
\cv{
\begin{table*}[t]
  \centering
\begin{tabular}{@{}lrrrrrrrrrrrr@{}}
\toprule
cuber & \multicolumn{4}{c}{Kochen-Specker} & \multicolumn{4}{c}{Murty-Simon} & \multicolumn{4}{c}{$\Delta$-free} \\
      & \multicolumn{2}{c}{sum} & \multicolumn{2}{c}{$\max$} & \multicolumn{2}{c}{sum} & \multicolumn{2}{c}{$\max$} & \multicolumn{2}{c}{sum} & \multicolumn{2}{c}{$\max$} \\
      & \default & \tuned & \default & \tuned & \default & \tuned & \default & \tuned & \default & \tuned & \default & \tuned \\
\midrule
\texttt{CDCL} & $510.4$ & $221.7$ & $9.1$ & $2.4$ & $112.4$ & $116.8$ & $0.1$ & $0.1$ & $601.6$ & $292.8$ & $6.6$ & $3.3$ \\
\texttt{CaDiCaL-LA-E} & $836.0$ & $290.3$ & $17.7$ & $7.9$ & $60.8$ & $56.0$ & $0.2$ & $0.2$ & $657.1$ & $260.3$ & $16.9$ & $7.3$ \\
\texttt{CaDiCaL-LA-}$\forall$ & $858.4$ & $296.5$ & $17.4$ & $3.4$ & $78.3$ & $71.7$ & $0.3$ & $0.2$ & $570.7$ & $219.7$ & $16.3$ & $5.4$ \\
\texttt{march} & $\mathbf{371.2}$ & $\mathbf{149.4}$ & $\mathbf{2.5}$ & $\mathbf{0.9}$ & $\mathbf{33.5}$ & $\mathbf{35.6}$ & $\mathbf{0.1}$ & $\mathbf{0.1}$ & $\mathbf{258.6}$ & $\mathbf{172.2}$ & $\mathbf{1.2}$ & $\mathbf{0.9}$ \\
\bottomrule
\end{tabular}
\caption{Total solving time (sum) and time to solve the hardest cube $\left( \max \right)$ in CPU hours for cubes produced for test instances by different cubers (rows) and solved by default or hyper-parameter-tuned SMS.The best cuber for each column is in boldface.}
\label{table:default-vs-tuned}
 	\vspace{-2ex} 
\end{table*}
}

In Figure~\ref{fig:histo} we show more detailed information about the distribution of running times over individual cubes.
For each benchmark problem we show 4 plots: the rows differ in the cuber, the columns differ in the conquering solver.
We compare default and best cuber, and also default and best configuration of CaDiCaL.
The plots show the total running time for cubes, grouped by individual running time.
The total area of each plot corresponds to total solving time, and we can see that as we move right (within the group of four plots corresponding to one benchmark problem), the area shrinks and shifts to the left.
This means that the total solving time is reduced, and is concentrated more into easier cubes, meaning that parallelization is going to be more effective.

Overall, from both Table~\ref{table:default-vs-tuned} and Figure~\ref{fig:histo} we can see that both cubing and algorithm configuration are very effective for these problems, and that our prerun-based pipeline is effective for cubing with SMS, in particular with march.

\section{Conclusion}\label{sec:conc}

Our experimental evaluation of cube-and-conquer pipelines with SMS revealed several key insights. March\_cu emerged as the best-performing cuber in our setup for handling propagator-generated clauses, effectively balancing global search coverage with local constraint propagation. Our runtime analysis showed that harder instances benefit disproportionately from parameter tuning, suggesting default solver heuristics may be suboptimal for specialized subproblems. This relationship between cube characteristics and solving strategies merits further investigation.

Several promising directions emerge from these findings. First, applying our LLM-guided scoring optimization to march\_cu could yield additional improvements. Second, while march\_cu generates well-balanced subproblems, our analysis suggests potential for specialized splitting strategies that account for propagator behavior. Finally, our success with automatic configuration raises interesting questions about parameter space structure, as the preliminary analysis indicates cube clusters that might benefit from targeted approaches. Our methodology demonstrates how carefully handling dynamically learned constraints can substantially improve parallel SAT-solving performance.


\begin{thebibliography}{}

\bibitem[\protect\citeauthoryear{Biere \bgroup \em et al.\egroup
  }{2024}]{BiereFallerFazekasFleuryFroleyks24}
Armin Biere, Tobias Faller, Katalin Fazekas, Mathias Fleury, Nils Froleyks, and
  Florian Pollitt.
\newblock {CaDiCaL 2.0}.
\newblock In Arie Gurfinkel and Vijay Ganesh, editors, {\em Computer Aided
  Verification - 36th International Conference, {CAV} 2024, Montreal, QC,
  Canada, July 24-27, 2024, Proceedings, Part {I}}, volume 14681 of {\em
  Lecture Notes in Computer Science}, pages 133--152. Springer, 2024.
\newblock DOI: \url{https://doi.org/10.1007/978-3-031-65627-9\_7}

\bibitem[\protect\citeauthoryear{Budroni \bgroup \em et al.\egroup
  }{2022}]{BudroniEtal22}
Costantino Budroni, Ad\'{a}n Cabello, Otfried G\"{u}hne, Matthias Kleinmann,
  and Jan-{\r A}ke Larsson.
\newblock {Kochen-Specker} contextuality.
\newblock {\em Rev. Mod. Phys.}, 94:045007, 2022.
\newblock DOI: \url{https://doi.org/10.1103/RevModPhys.94.045007}

\bibitem[\protect\citeauthoryear{Caccetta and
  H\"{a}ggkvist}{1979}]{CaccettaHaggkvist79}
Louis Caccetta and Roland H\"{a}ggkvist.
\newblock On diameter critical graphs.
\newblock {\em Discrete Math.}, 28(3):223--229, 1979.
\newblock DOI: \url{https://doi.org/10.1016/0012-365X(79)90129-8}

\bibitem[\protect\citeauthoryear{Chen and F\"{u}redi}{2005}]{ChenFuredy05}
Ya-Chen Chen and Zolt\'{a}n F\"{u}redi.
\newblock Minimum vertex-diameter-2-critical graphs.
\newblock {\em J. Graph Theory}, 50(4):293--315, 2005.
\newblock DOI: \url{https://doi.org/10.1002/jgt.20111}

\bibitem[\protect\citeauthoryear{Codish \bgroup \em et al.\egroup
  }{2019}]{CodishMillerProsserStuckey19}
Michael Codish, Alice Miller, Patrick Prosser, and Peter~J. Stuckey.
\newblock Constraints for symmetry breaking in graph representation.
\newblock {\em Constraints}, 24(1):1--24, 2019.
\newblock DOI: \url{https://doi.org/10.1007/s10601-018-9294-5}

\bibitem[\protect\citeauthoryear{Crawford \bgroup \em et al.\egroup
  }{1996}]{CrawfordGinsbergLuksRoy96}
James~M. Crawford, Matthew~L. Ginsberg, Eugene~M. Luks, and Amitabha Roy.
\newblock Symmetry-breaking predicates for search problems.
\newblock In Stuart C.~Shapiro Luigia Carlucci~Aiello, Jon~Doyle, editor, {\em
  Proceedings of the Fifth International Conference on Principles of Knowledge
  Representation and Reasoning (KR'96), Cambridge, Massachusetts, USA, November
  5-8, 1996}, pages 148--159. Morgan Kaufmann, 1996.

\bibitem[\protect\citeauthoryear{Dailly \bgroup \em et al.\egroup
  }{2019}]{DaillyFoucaudHansberg19}
Antoine Dailly, Florent Foucaud, and Adriana Hansberg.
\newblock Strengthening the {M}urty-{S}imon conjecture on diameter 2 critical
  graphs.
\newblock {\em Discrete Math.}, 342(11):3142--3159, 2019.
\newblock DOI: \url{https://doi.org/10.1016/j.disc.2019.06.023}

\bibitem[\protect\citeauthoryear{Erd\H{o}s and R\'{e}nyi}{1962}]{ErdosRenyi62}
P.~Erd\H{o}s and A.~R\'{e}nyi.
\newblock On a problem in the theory of graphs.
\newblock {\em A Magyar Tudom{\'a}nyos Akad{\'e}mia Matematikai Kutat{\'o}
  Int{\'e}zet{\'e}nek K{\"o}zlem{\'e}nyei}, 7(4):623--641 (1963), 1962.

\bibitem[\protect\citeauthoryear{Erd\H{o}s}{1967}]{Erdos67}
P.~Erd\H{o}s.
\newblock Some remarks on chromatic graphs.
\newblock {\em Colloq. Math.}, 16:253--256, 1967.
\newblock DOI: \url{https://doi.org/10.4064/cm-16-1-253-256}

\bibitem[\protect\citeauthoryear{Fan}{1987}]{Fan87}
Genghua Fan.
\newblock On diameter 2-critical graphs.
\newblock {\em Discret. Math.}, 67(3):235--240, 1987.
\newblock DOI: \url{https://doi.org/10.1016/0012-365X(87)90174-9}

\bibitem[\protect\citeauthoryear{Fazekas \bgroup \em et al.\egroup
  }{2023}]{FazekasNPKSB23}
Katalin Fazekas, Aina Niemetz, Mathias Preiner, Markus Kirchweger, Stefan
  Szeider, and Armin Biere.
\newblock {IPASIR-UP}: User propagators for {CDCL}.
\newblock In Meena Mahajan and Friedrich Slivovsky, editors, {\em The 26th
  International Conference on Theory and Applications of Satisfiability Testing
  (SAT 2023), July 04-08, 2023, Alghero, Italy}, LIPIcs. Schloss Dagstuhl -
  Leibniz-Zentrum f{\"{u}}r Informatik, 2023.

\bibitem[\protect\citeauthoryear{Fazekas \bgroup \em et al.\egroup
  }{2024}]{FazekasNPKSB24}
Katalin Fazekas, Aina Niemetz, Mathias Preiner, Markus Kirchweger, Stefan
  Szeider, and Armin Biere.
\newblock Satisfiability modulo user propagators.
\newblock {\em J. Artif. Intell. Res.}, 81:989--1017, 2024.

\bibitem[\protect\citeauthoryear{Fichte \bgroup \em et al.\egroup
  }{2023a}]{FichteLeberreHecherSzeider23}
Johannes~K. Fichte, Daniel~Le Berre, Markus Hecher, and Stefan Szeider.
\newblock The silent (r)evolution of {SAT}.
\newblock {\em Communications of the ACM}, 66(6):64--72, June 2023.
\newblock DOI: \url{https://doi.org/10.1145/3560469}

\bibitem[\protect\citeauthoryear{Fichte \bgroup \em et al.\egroup
  }{2023b}]{FichteHLS23}
Johannes~K. Fichte, Markus Hecher, Daniel~Le Berre, and Stefan Szeider.
\newblock The silent (r)evolution of {SAT}.
\newblock {\em Communications of the ACM}, 66(6):64--72, June 2023.
\newblock DOI: \url{https://doi.org/10.1145/3560469}

\bibitem[\protect\citeauthoryear{Goedgebeur}{2020}]{Goedgebeur20}
Jan Goedgebeur.
\newblock On minimal triangle-free 6-chromatic graphs.
\newblock {\em J. Graph Theory}, 93(1):34--48, 2020.
\newblock DOI: \url{https://doi.org/10.1002/jgt.22467}

\bibitem[\protect\citeauthoryear{Haynes \bgroup \em et al.\egroup
  }{2015}]{HaynesEtal15}
Teresa~W. Haynes, Michael~A. Henning, Lucas~C. van~der Merwe, and Anders Yeo.
\newblock Progress on the {M}urty-{S}imon {C}onjecture on diameter-2 critical
  graphs: a survey.
\newblock {\em J. Comb. Optim.}, 30(3):579--595, 2015.
\newblock DOI: \url{https://doi.org/10.1007/s10878-013-9651-7}

\bibitem[\protect\citeauthoryear{Heule and van Maaren}{2006}]{HeuleM06}
Marijn Heule and Hans van Maaren.
\newblock March{\_}dl: Adding adaptive heuristics and a new branching strategy.
\newblock {\em J. Satisf. Boolean Model. Comput.}, 2(1-4):47--59, 2006.
\newblock DOI: \url{https://doi.org/10.3233/SAT190016}

\bibitem[\protect\citeauthoryear{Heule and van Maaren}{2009}]{HeuleM09}
Marijn Heule and Hans van Maaren.
\newblock Look-ahead based {SAT} solvers.
\newblock In Armin Biere, Marijn Heule, Hans van Maaren, and Toby Walsh,
  editors, {\em Handbook of Satisfiability}, volume 185 of {\em Frontiers in
  Artificial Intelligence and Applications}, pages 155--184. {IOS} Press, 2009.
\newblock DOI: \url{https://doi.org/10.3233/978-1-58603-929-5-155}

\bibitem[\protect\citeauthoryear{Heule \bgroup \em et al.\egroup
  }{2004}]{HeuleDZM04}
Marijn Heule, Mark Dufour, Joris~E. van Zwieten, and Hans van Maaren.
\newblock March{\_}eq: Implementing additional reasoning into an efficient
  look-ahead {SAT} solver.
\newblock In Holger~H. Hoos and David~G. Mitchell, editors, {\em Theory and
  Applications of Satisfiability Testing, 7th International Conference, {SAT}
  2004, Vancouver, BC, Canada, May 10-13, 2004, Revised Selected Papers},
  volume 3542 of {\em Lecture Notes in Computer Science}, pages 345--359.
  Springer, 2004.
\newblock DOI: \url{https://doi.org/10.1007/11527695\_26}

\bibitem[\protect\citeauthoryear{Heule \bgroup \em et al.\egroup
  }{2016}]{HeuleKullmannMarek16}
Marijn J.~H. Heule, Oliver Kullmann, and Victor~W. Marek.
\newblock Solving and verifying the boolean pythagorean triples problem via
  cube-and-conquer.
\newblock In Nadia Creignou and Daniel~Le Berre, editors, {\em Theory and
  Applications of Satisfiability Testing - {SAT} 2016 - 19th International
  Conference, Bordeaux, France, July 5-8, 2016, Proceedings}, volume 9710 of
  {\em Lecture Notes in Computer Science}, pages 228--245. Springer Verlag,
  2016.

\bibitem[\protect\citeauthoryear{Heule \bgroup \em et al.\egroup
  }{2018}]{HeuleKullmannBiere18}
Marijn J.~H. Heule, Oliver Kullmann, and Armin Biere.
\newblock Cube-and-conquer for satisfiability.
\newblock In Youssef Hamadi and Lakhdar Sais, editors, {\em Handbook of
  Parallel Constraint Reasoning}, pages 31--59. Springer, 2018.
\newblock DOI: \url{https://doi.org/10.1007/978-3-319-63516-3_2}

\bibitem[\protect\citeauthoryear{Janota \bgroup \em et al.\egroup
  }{2025}]{JanotaKirchwegerPeitlSzeider25}
Mikol{\'{a}}{\v{s}} Janota, Markus Kirchweger, Tom{\'{a}}{\v{s}} Peitl, and
  Stefan Szeider.
\newblock Breaking symmetries in quantified graph search: A comparative study.
\newblock In {\em {AAAI} 2025: The 39th Annual AAAI Conference on Artificial
  Intelligence}, 2025.

\bibitem[\protect\citeauthoryear{Kirchweger and
  Szeider}{2021}]{KirchwegerSzeider21}
Markus Kirchweger and Stefan Szeider.
\newblock {SAT} modulo symmetries for graph generation.
\newblock In {\em 27th International Conference on Principles and Practice of
  Constraint Programming (CP 2021)}, LIPIcs, page 39:1–39:17. Dagstuhl, 2021.
\newblock DOI: \url{https://doi.org/10.4230/LIPIcs.CP.2021.34}

\bibitem[\protect\citeauthoryear{Kirchweger and
  Szeider}{2024}]{KirchwegerSzeider24}
Markus Kirchweger and Stefan Szeider.
\newblock {SAT} modulo symmetries for graph generation and enumeration.
\newblock {\em {ACM} Trans. Comput. Log.}, 25(3):1--30, 2024.
\newblock DOI: \url{https://doi.org/10.1145/3670405}

\bibitem[\protect\citeauthoryear{Kirchweger \bgroup \em et al.\egroup
  }{2022}]{KirchwegerScheucherSzeider22}
Markus Kirchweger, Manfred Scheucher, and Stefan Szeider.
\newblock A {SAT} attack on {Rota's Basis Conjecture}.
\newblock In {\em Theory and Applications of Satisfiability Testing - {SAT}
  2022 - 25th International Conference, Haifa, Israel, August 2-5, 2022,
  Proceedings}, 2022.
\newblock DOI: \url{https://doi.org/10.4230/LIPIcs.SAT.2022.4}

\bibitem[\protect\citeauthoryear{Kirchweger \bgroup \em et al.\egroup
  }{2023a}]{KirchwegerPeitlSzeider23}
Markus Kirchweger, Tom{\'{a}}s Peitl, and Stefan Szeider.
\newblock Co-certificate learning with {SAT} modulo symmetries.
\newblock In {\em Proceedings of the Thirty-Second International Joint
  Conference on Artificial Intelligence, {IJCAI} 2023, 19th-25th August 2023,
  Macao, SAR, China}, pages 1944--1953. ijcai.org, 2023.
\newblock DOI: \url{https://doi.org/10.24963/IJCAI.2023/216}

\bibitem[\protect\citeauthoryear{Kirchweger \bgroup \em et al.\egroup
  }{2023b}]{KirchwegerPeitlSzeider23b}
Markus Kirchweger, Tom{\'{a}}s Peitl, and Stefan Szeider.
\newblock A {SAT} solver's opinion on the {Erd{\H{o}}s}-{Faber}-{Lov{\'{a}}sz}
  conjecture.
\newblock In {\em 26th International Conference on Theory and Applications of
  Satisfiability Testing, {SAT} 2023, July 4-8, 2023, Alghero, Italy}, volume
  271 of {\em LIPIcs}, pages 13:1--13:17. Schloss Dagstuhl - Leibniz-Zentrum
  f{\"{u}}r Informatik, 2023.
\newblock DOI: \url{https://doi.org/10.4230/LIPIcs.SAT.2023.13}

\bibitem[\protect\citeauthoryear{Kirchweger \bgroup \em et al.\egroup
  }{2023c}]{KirchwegerScheucherSzeider23}
Markus Kirchweger, Manfred Scheucher, and Stefan Szeider.
\newblock {SAT}-based generation of planar graphs.
\newblock In Meena Mahajan and Friedrich Slivovsky, editors, {\em The 26th
  International Conference on Theory and Applications of Satisfiability Testing
  (SAT 2023), July 04-08, 2023, Alghero, Italy}, LIPIcs. Schloss Dagstuhl -
  Leibniz-Zentrum f{\"{u}}r Informatik, 2023.
\newblock DOI: \url{https://doi.org/10.4230/LIPICS.SAT.2023.14}

\bibitem[\protect\citeauthoryear{Kirchweger \bgroup \em et al.\egroup
  }{2025}]{PeitlKSX25}
Markus Kirchweger, Tom{\'{a}}s Peitl, Stefan Szeider, and Hai Xia.
\newblock Supplementary material for smart cubing for graph search: A
  comparative study.
\newblock Online, Zenodo, 2025.
\newblock DOI: \url{https://doi.org/10.5281/zenodo.14746543}

\bibitem[\protect\citeauthoryear{Kochen and Specker}{1967}]{KochenSpecker67}
Simon Kochen and Ernst Specker.
\newblock The problem of hidden variables in quantum mechanics.
\newblock {\em J. Math. Mech.}, 17(1):59--87, 1967.

\bibitem[\protect\citeauthoryear{Li \bgroup \em et al.\egroup
  }{2024}]{LiBrightGanesh24}
Zhengyu Li, Curtis Bright, and Vijay Ganesh.
\newblock A sat solver + computer algebra attack on the minimum
  kochen–specker problem.
\newblock In Kate Larson, editor, {\em Proceedings of the Thirty-Third
  International Joint Conference on Artificial Intelligence, {IJCAI-24}}, pages
  1898--1906. International Joint Conferences on Artificial Intelligence
  Organization, 8 2024.
\newblock Main Track.
\newblock DOI: \url{https://doi.org/10.24963/ijcai.2024/210}

\bibitem[\protect\citeauthoryear{Lindauer \bgroup \em et al.\egroup
  }{2022}]{SMAC3}
Marius Lindauer, Katharina Eggensperger, Matthias Feurer, André Biedenkapp,
  Difan Deng, Carolin Benjamins, Tim Ruhkopf, René Sass, and Frank Hutter.
\newblock Smac3: A versatile bayesian optimization package for hyperparameter
  optimization.
\newblock {\em Journal of Machine Learning Research}, 23(54):1--9, 2022.
\newblock URL: \url{http://jmlr.org/papers/v23/21-0888.html}

\bibitem[\protect\citeauthoryear{Loh and Ma}{2016}]{LohMa16}
Po-Shen Loh and Jie Ma.
\newblock Diameter critical graphs.
\newblock {\em J. Combin. Theory Ser. B}, 117:34--58, 2016.
\newblock DOI: \url{https://doi.org/10.1016/j.jctb.2015.11.004}

\bibitem[\protect\citeauthoryear{Mantel}{1907}]{Mantel07}
W.~Mantel.
\newblock Problem 28.
\newblock {\em Wiskundige Opgaven}, 10:60--61, 1907.

\bibitem[\protect\citeauthoryear{Marques{-}Silva \bgroup \em et al.\egroup
  }{2021}]{MarquessilvaLynceMalik21}
Jo{\~{a}}o Marques{-}Silva, In{\^{e}}s Lynce, and Sharad Malik.
\newblock Conflict-driven clause learning {SAT} solvers.
\newblock In Armin Biere, Marijn Heule, Hans van Maaren, and Toby Walsh,
  editors, {\em Handbook of Satisfiability - Second Edition}, volume 336 of
  {\em Frontiers in Artificial Intelligence and Applications}, pages 133--182.
  {IOS} Press, 2021.
\newblock DOI: \url{https://doi.org/10.3233/FAIA200987}

\bibitem[\protect\citeauthoryear{McKay and Piperno}{2014}]{McKayPiperno14}
Brendan~D. McKay and Adolfo Piperno.
\newblock Practical graph isomorphism, {II}.
\newblock {\em J. Symbolic Comput.}, 60:94--112, 2014.
\newblock DOI: \url{https://doi.org/10.1016/j.jsc.2013.09.003}

\bibitem[\protect\citeauthoryear{Mycielski}{1955}]{Mycielski55}
Jan Mycielski.
\newblock Sur le coloriage des graphs.
\newblock {\em Colloquium Mathematicae}, 3(2):161--162, 1955.

\bibitem[\protect\citeauthoryear{Peres}{1991}]{Peres91}
A~Peres.
\newblock Two simple proofs of the {Kochen-Specker} theorem.
\newblock {\em J. Phys. A Math. Theor.}, 24(4):L175, feb 1991.
\newblock DOI: \url{https://doi.org/10.1088/0305-4470/24/4/003}

\bibitem[\protect\citeauthoryear{Radosavljevi\'{c} and
  \v{Z}ivkovi\'{c}}{2020}]{RadosavljevicZivkovic20}
Jovan Radosavljevi\'{c} and Miodrag \v{Z}ivkovi\'{c}.
\newblock The list of diameter-2-critical graphs with at most 10 nodes.
\newblock {\em IPSI Transactions on Advanced Research}, 16(1):1--5, January
  2020.
\newblock \url{http://ipsitransactions.org/journals/papers/tar/2020jan/p9.pdf}.

\bibitem[\protect\citeauthoryear{Zhang and Szeider}{2023}]{ZhangSzeider23}
Tianwei Zhang and Stefan Szeider.
\newblock Searching for smallest universal graphs and tournaments with {SAT}.
\newblock In {\em 29th International Conference on Principles and Practice of
  Constraint Programming, {CP} 2023, August 27-31, 2023, Toronto, Canada},
  volume 280 of {\em LIPIcs}, pages 39:1--39:20. Schloss Dagstuhl -
  Leibniz-Zentrum f{\"{u}}r Informatik, 2023.
\newblock DOI: \url{https://doi.org/10.4230/LIPICS.CP.2023.39}

\bibitem[\protect\citeauthoryear{Zhang \bgroup \em et al.\egroup
  }{2024}]{ZhangPeitlSzeider24}
Tianwei Zhang, Tom{\'{a}}s Peitl, and Stefan Szeider.
\newblock Small unsatisfiable k-cnfs with bounded literal occurrence.
\newblock In Supratik Chakraborty and Jie{-}Hong~Roland Jiang, editors, {\em
  27th International Conference on Theory and Applications of Satisfiability
  Testing, {SAT} 2024, August 21-24, 2024, Pune, India}, volume 305 of {\em
  LIPIcs}, pages 31:1--31:22. Schloss Dagstuhl - Leibniz-Zentrum f{\"{u}}r
  Informatik, 2024.
\newblock DOI: \url{https://doi.org/10.4230/LIPICS.SAT.2024.31}

\end{thebibliography}
\end{document}